# On Reasoning with Ambiguities

Uwe Reyle
Institute for Computational Linguistics
University of Stuttgart
Azenbergstr.12, D–70174 Stuttgart, Germany
e-mail: uwe@ims.uni-stuttgart.de

## Abstract

The paper adresses the problem of reasoning with ambiguities. Semantic representations are presented that leave scope relations between quantifiers and/or other operators unspecified. Truth conditions are provided for these representations and different consequence relations are judged on the basis of intuitive correctness. Finally inference patterns are presented that operate directly on these underspecified structures, i.e. do not rely on any translation into the set of their disambiguations.

## 1 Introduction

Whenever we hear a sentence or read a text we build up mental representations in which some aspects of the meaning of the sentence or text are left underspecified. And if we accept what we have heard or read as true, then we will use these underspecified representations as premises for arguments. The challenge is, therefore, to equip underspecified semantic representations with well-defined truth conditions and to formulate inference patterns for these representations that follow the arguments that we judge as intuitively correct. Several proposals exist for the definition of the language, but only very few authors have addressed the problem of defining a logic of ambiguous reasoning.

[8] considers lexical ambiguities and investigates structural properties of a number of consequence relations based on an abstract notion of coherency. It is not clear, however, how this approach could be extended to other kinds of ambiguities, especially quantifier scope ambiguities and ambiguities triggered by plural NPs. [1], [7] and [6] deal with ambiguities of the latter kind. They give construction rules and define truth conditions according to which an underspecified representation of an ambiguous sentence is true if one of its disambiguations is. The problem of reasoning is adressed only in [5] and [7]. [5]'s inference schemata yield a very weak logic only; and [7]'s deductive component is too strong. Being weak and strong depends of course on the underlying consequence relation. Neither [5] nor [7] make any attempt to systematically derive the consequence relation that holds for reasoning with ambiguities on the basis of an empirical discussion of intuitively valid arguments.

The present paper starts out with such a discussion in Section 2. Section 3 gives a brief introduction to the theory of UDRSs. It gives a sketch of the principles to construct UDRSs and shows how scope ambiguities of quantifiers and negation are represented in an underspecified way. As the rules of inference presented in [7] turn out to be sound also with respect to the consequence relation defined in Section 2 these rules (for the fragment without disjunction) will be discussed only briefly in Section 4. The change in the deduction system that is imposed by the new consequence relation comes with the rules of proof. Section 5 shows that it is no longer possible to use rules like Conditionalisation or Reductio ad Absurdum when we deal with real ambiguities in the goal. An alternative set of rules is presented in Section 6.

## 2 Consequence Relations

In this section we will discuss some sample arguments containing ambiguous expressions in the data as well as in the goal. We consider three kinds of ambiguities: lexical ambiguities, quantifier scope ambiguities, and ambiguities with respect to distributive/collective readings of plural noun phrases. The discussion of the arguments will show that the meaning of ambiguous sentences not only depends on the set of its disambiguations. Their meanings also depend on the context, especially on other occurrences of ambiguities. Each disambiguation of an ambiguous sentence may be correlated to disambiguations of other ambiguous sentences such that the choice of the first disambiguation also determines the choice of the latter ones, and vice versa. Thus the representation of ambiguities requires some means to implement these correlations.

To see that this is indeed the case let us start discussing some consequence relations that come to mind when dealing with ambiguous reasoning. The first one we will consider is the one that allows to derive

a(n ambiguous) conclusion $\gamma$ from a set of (ambiguous) premisses $\Gamma$ if some disambiguation of $\gamma$ follows from all readings of $\Gamma$. Assuming that $\delta$ and $\delta'$ are operators mapping a set of ambiguous representations $\alpha$ onto one of its disambiguations $\alpha^\delta$ or $\alpha^{\delta'}$ we may represent this by.

(1) $\qquad \forall \delta \exists \delta' (\Gamma^\delta \models \gamma^{\delta'})$.

Obviously (1) is the relation we get if we interpret ambiguities as being equivalent to the disjunctions of their readings. To interpret ambiguities in this way is, however, not correct. For ambiguities in the goal this is witnessed by (2).

(2) $\models$ Everybody slept or everybody didn't sleep.

Intuitively (2) is contingent, but would – according to the relation in (1) – be classified as a tautology. In this case the consequence relation in (3) gives the correct result and therefore seems to be preferable.

(3) $\qquad \forall \delta \forall \delta' (\Gamma^\delta \models \gamma^{\delta'})$

But there is another problem with (3). It does not fulfill Reflexivity, which (1) does.

**Reflexivity** $\quad \Gamma \models \psi$, if $\psi \in \Gamma$

To do justice to both, the examples in (2) and Reflexivity, we would have to interpret ambiguous sentences in the data also as conjunctions of their readings, i.e. accept (4) as consequence relation.

(4) $\qquad \exists \delta' \exists \delta (\Gamma^\delta \models \gamma^{\delta'})$

But this again contradicts intuitions. (4) would support the inferences in (5), which are intuitively not correct.

(5)
 a. There is a big plant in front of my house.
 $\models$ There is a big building in front of my house.
 b. Everybody didn't sleep. $\models$ Everybody was awake.
 c. Three boys got £10. $\models$ Three boys got £10 each.

Given the examples in (5) we are back to (1) and may think that ambiguities in the data are interpreted as disjunctions of their readings. But irrespective of the incompatibility with Reflexivity this picture cannot be correct either, because it destroys the intuitively valid inference in (6).

(6) If the students get £10 then they buy books.
 The students get £10. $\models$ They buy books.

This example shows that disambiguation is not an operation $\delta$ that takes (a set of) isolated sentences. Ambiguous sentences of the same type have to be disambiguated simultaneously.[1] Thus the meaning of the premise of (6) is given by (7b) not by (7a), where $a_1$ represents the first and $a_2$ the second reading of the second sentence of (6).

(7)
 a. $((a_1 \to b) \lor (a_2 \to b)) \land (a_1 \lor a_2)$
 b. $((a_1 \to b) \land a_1) \lor ((a_2 \to b) \land a_2)$

We will call sentence representations that have to be disambiguated simultaneously *correlated ambiguities*. The correlation may be expressed by coindexing. Any disambiguation $\delta$ that simultaneously disambiguates a set of representations coindexed with $i$ is a disambiguation that *respects $i$*, in symbols $\delta^i$. A disambiguation $\delta$ that respects all indices of a given set $I$ is said to *respect $I$*, written $\delta^I$. Let $I$ be a set of indices, then the consequence relation we assume to underly ambiguous reasoning is given in (8)

(8) $\qquad \forall \delta^I (\Gamma^{\delta^I} \models \gamma^{\delta^I})$

The general picture we will follow in this paper is the following. We assume that a set of representations $\Gamma$ represents the mental state of a reasoning agent R. $\Gamma$ contains underspecified representations. Correlations between elements of $\Gamma$ indicate that they share possible ways of disambiguation. Suppose $\gamma$ is only implicitly contained in $\Gamma$. Then R may infer it from $\Gamma$ and make it explicit by adding it to its mental state. This process determines the consequence relation relative to which we develop our inference patterns. That means we do not consider the case where R is asked some query $\gamma$ by another person B. The additional problem in this case consists in the array of possibilities to establish correlations between B's query and R's data, and must be adressed within a proper theory of dialogue.

Consider the following examples. The data contains two clauses. The first one is ambiguous, but not in the context of the second.

(9)
 a. Every pitcher was broken. They had lost.
  $\models$ Every pitcher was broken.
 b. Everybody didn't sleep. John was awake.
  $\models$ Everybody didn't sleep.
 c. John and Mary bought a house.
  It was completely delapidated.
  $\models$ John and Mary bought a house.

If the inference is now seen as the result of R's task to make the first sentence explicit (which of course is trivial here), then the goal will not be ambiguous, because it simply is another occurrence of the representation in the data, and, therefore, will carry the same correlation index. In the second case, i.e. the case where the goal results from R's processing some external input, there is no guarantee for such a correlation. R might consider the goal as ambiguous, and hence will not accept it as a consequence. (B might after all have had in mind just that reading of the sentence that is not part of R's knowledge.)

---

[1] We will not give a classification or definition of *ambiguities of the same type* here. Three major classes will consist of lexical ambiguities, ambiguities with respect to distributive/collective readings of plural noun phrases, and quantifier scope ambiguities. As regards the last type we assume on the one hand that only sentences with the same argument structure and the same set of readings can be of the same type. More precisely, if two sentences are of the same type with respect to quantifier scope ambiguities, then the labels of their UDRS's must be ordered isomorphically. On the other hand two sentences may carry an ambiguity of the same type if one results from the other by applying Detachment to a universally quantified NP (see Section 4).

We will distinguish between these two situations by requiring the provability relation to respect indices. The rule of direct proof will then be an instance of Reflexivity: $\Gamma \vdash \gamma_i$ if $\gamma_i \in \Gamma$.

## 3 A short introduction to UDRSs

The base for unscoped representations proposed in [7] is the separation of information about the structure of a particular semantic form and of the content of the information bits the semantic form combines. In case the semantic form is given by a DRS its structure is given by the hierarchy of subDRSs, that is determined by $\Rightarrow, \neg, \vee$ and $\Diamond$. We will represent this hierarchy explicitly by the subordination relation $\leq$. The semantic content of a DRS consists of the set of its discourse referents and its conditions. To be more precise, we express the structural information by a language with one predicate $\leq$ that relates individual constants $l$, called *labels*. The constants are names for DRS's. $\leq$ corresponds to the subordination relation between them, i.e. the set of labels with $\leq$ is a upper semilattice with one-element (denoted by $l_\top$). Let us consider the DRSs (11) and (12) representing the two readings of (10).

(10)  Everybody didn't pay attention.

(11) 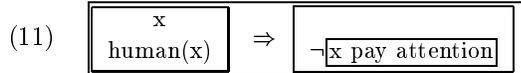

(12) 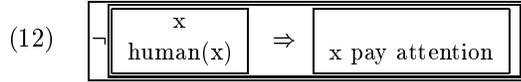

The following representations make the distinction between structure and content more explicit. The subordination relation $\leq$ is read from bottom to top.

(13) 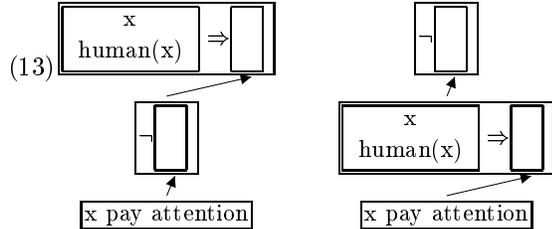

Having achieved this separation we are able to represent the structure that is common to both, (11) and (12), by (14).

(14) 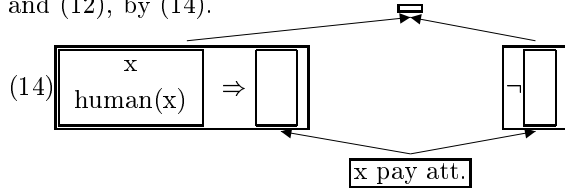

(14) is already the UDRS that represents (10) with scope relationships left unresolved. We call the nodes of such graphs UDRS-*components*. Each UDRS-component consists of a labelled DRS and two functions *scope* and *res*, which map labels of UDRS-components to the labels of their scope and restrictor, respectively. DRS-conditions are of the form $\langle Q, l_{i1}, l_{i2} \rangle$, with quantifier $Q$, restrictor $l_{i1}$ and scope $l_{i2}$, of the form $l_{i1} \Rightarrow l_{i2}$, or of the form $l_i : \neg l_{i1}$. A UDRS is a set of UDRS-components together with a partial order ORD of its labels.

If we make (some) labels explicit we may represent (14) as in (15).

(15) $\langle l_\top : \langle l_1 : \boxed{\begin{array}{c} x \\ \end{array}} \Rightarrow \boxed{\phantom{xx}}, l_2 : \neg \boxed{\phantom{xx}}, l_3 : \boxed{\text{x pay att.}} \rangle, \text{ORD} \rangle$

If $ORD$ in (15) is given as $\{l_2 \leq scope(l_1), l_3 \leq scope(l_2)\}$ then (15) is equivalent to (11), and in case $ORD$ is $\{l_1 \leq scope(l_2), l_3 \leq scope(l_1)\}$ we get a description of (12). If $ORD$ is $\{l_3 \leq scope(l_1), l_3 \leq scope(l_2)\}$ then (15) represents (14), because it only contains the information common to both, (11) and (12).

In any case ORD lists only the subordination relations that are neither implicitly contained in the partial order nor determined by complex UDRS-conditions. This means that (15) implicitly contains the information that, e.g., $res(l_2) \leq l_\top$, and also that $res(l_2) \leq l_2$, $res(l_1) \leq l_\top$ and $scope(l_1) \leq l_\top$.

In this paper we consider the fragment of UDRSs without disjunction. For reason of space we cannot consider problems that arise when indefinites occurring in subordinate clauses are interpreted specifically.[2] We will, therefore assume that indefinites behave like generalized quantifers in that their scope is clause bounded too, i.e. require $l \leq l'$ for all $i$ in clause (ii.c) of the following definition.

**Definition 1:**

(i) $\langle l : \langle U_K, C'_K \cup C''_K \rangle, res(l), scope(l), ORD_l \rangle$ is a *UDRS-component*, if $\langle U_K, C'_K \rangle$ is a DRS containing standard DRS-conditions only, and $C''_K$ is one of the following sets of *labelled* DRS-conditions, where $K_1$ and $K_2$ are standard DRSs, $Q_x$ is a generalized quantification over $x$, and $l'$ is the upper bound of a (subordinate) UDRS-clause $\langle l' : \langle \gamma_0, ..., \gamma_n \rangle, ORD'_l \rangle$ (defined below).

(a) $\{\}$, or $\{sub(l')\}$

(b) $\{l_1 \Rightarrow l_2, l_1 : K_1, l_2 : K_2\}$, or
    $\{l_1 \Rightarrow l_2, l_1 : K_1, l_2 : K_2, l_1 : sub(l')\}$

(c) $\{\langle Q_x l_1, l_2 \rangle, l_1 : K_1, l_2 : K_2\}$, or
    $\{\langle Q_x l_1, l_2 \rangle, l_1 : K_1, l_2 : K_2, l_1 : sub(l')\}\}$[3]

(d) $\{\neg l_1, l_1 : K_1\}$

If $C''_K \neq \{\}$ then $l_1 \Rightarrow l_2$, $\langle Q_x l_1, l_2 \rangle$, or $\neg l_1$ is called *distinguished condition* of $K$, referred to by $l : \gamma$.

*res* and *scope* are functions on the set of labels, and $ORD_l$ is a partial order of labels. $res(l)$, $scope(l)$, and $ORD_l$ are subject to the following restrictions:

---

[2] These problems are discussed extensively in [7] and the solution given there can be taken over to the rules presented here.

[3] Whenever convenient we will simply use implicative conditions of the form $l_1 \Rightarrow l_2$, to represent universally quantified NPs (instead of their generalized quantifier representation $\langle every, l_1, l_2 \rangle$).

(a) ($\alpha$) If $\neg l_1 \in C''_K$, then
$res(l) = scope(l) = l_1$ and $l_1 < l \in ORD_l$.[4]

($\beta$) If $\langle \Rightarrow, l_1, l_2 \rangle \in C''_K$, or $Q_x l_1, l_2 \in C''_K$, then
$res(l) = l_1$, $scope(l) = l_2$, and $l_1 < l$, $l_2 < l$, $l_1 \neq l_2 \in ORD_l$.

($\gamma$) Otherwise $res(l) = scope(l) = l$

(b) If $k{:}sub(l') \in C''_K$, then $l' < k \in ORD_l$ and $ORD_{l'} \subset ORD_l$.

(ii) A *UDRS-clause* is a pair $\langle l{:}\langle \gamma_0, ..., \gamma_n \rangle, ORD_l \rangle$, where $\gamma_i = \langle l_i{:}K_i, res(l_i), scope(l_i), ORD_{l_i} \rangle$, $0 \leq i \leq n$, are UDRS components, and $ORD_l$ contains all of the conditions in (a) to (c) and an arbitrary subset oif those in (d) and (e).

(a) $ORD_{l_i} \subset ORD_l$, for all $i$, $0 \leq i \leq n$

(b) $l_0 \leq scope(l_i) \in ORD_l$ for all $i$, $1 \leq i \leq n$

(c) $l_i \leq l \in ORD_l$ for all $i$, $1 \leq i \leq n$.

(d) $l_i \leq scope(l_j) \in ORD_l$, for some $i,j$ $1 \leq i,j \leq n$ such that ORD is a partial order.

For each $i$, $1 \leq i \leq n$, $l_i$ is called a *node*. $l$ is called *upper bound* and $l_0$ *lower bound* of the UDRS-clause. Lower bounds neither have distinguished conditions nor is there an $l'$ such that $l' < l$.

(iii) A *UDRS-database* is a set of UDRSs $\{\langle l^i_\top {:} \Gamma, ORD_{l^i_\top} \rangle\}_i$. A *UDRS-goal* is a UDRS.

For the fragment of this paper UDRS-components that contain distinguished conditions do not contain anything else, i.e. they consist of labelled DRSs $K$ for which $U_K = C'_K = \{\}$ if $C''_K \neq \{\}$. We assume that semantic values of verbs are associated with lower bounds of UDRS-clauses and NP-meanings with their other components. Then the definition of UDRSs ensures that[5]

(i) the verb is in the scope of each of its arguments, (clause (ii.b)),

(ii) the scope of proper quantifiers is clause bounded, (clause (ii.c))

For relative clauses the upper bound label $l'$ is subordinated to the label $l$ of its head noun (i.e. the restrictor of the NP containing the relative) by $l' < l$ (see (ii)). In the case of conditionals the upper bound label of subordinate clauses is set equal to the label of the antecedent/consequent of the implicative condition. The ordering of the set of labels of a UDRS builds an upper-semilattice with one-element $l_\top$. We assume that databases are constructed out of sequences $S_1, ..., S_n$ of sentences. Having a unique one-element $l^i_\top$ associated with each UDRS representing a sentence $S_i$ is to prevent any quantifier of $S_i$ to have scope over (parts of) any other sentence.

## 4 Rules of Inference

The four inference rules needed for the fragment without generalized quantifiers[6] and disjunction are non-empty universe (NeU), detachment (DET), ambiguity introduction (AI), and ambiguity elimination (DIFF). NeU allows to add any finite collection of discourse referents to a DRS universe. It reflects the assumption that there is of necessity one thing, i.e. that we consider only models with non-empty universes. DET is a generalization of modus ponens. It allows to add (a variant of) the consequent of an implication (or the scope of a universally quantified condition) to the DRS in which the condition occurs if the antecedent (restrictor) can be mapped to this DRS. AI allows one to add an ambiguous representation to the data, if the data already contains all of its disambiguations. And an application of DIFF reduces the set of readings of an underspecified representation in the presence of negations of some of its readings. The formulations of NeU, DET and DIFF needed for the consequence relation (8) defined in Section 2 of this paper are just refinements of the formulations needed for the consequence relation (1). As the latter case is extensively discussed in [7] and a precise and complete formulation of the rules is also given there we will restrict ourselves to the refinements needed to adapt these rules to the new consequence relation.

As there is nothing more to mention about NeU we start with DET. We first present a formulation of DET for DRSs. It is an extended formulation of standard DET as it allows for applications not only at the top level of a DRS but at levels of any depth. Correctness of this extension is shown in [4].

**DET** Suppose a DRS K contains a condition of the form $K_1 \Rightarrow K_2$ such that $K_1$ may be embedded into $\overline{K}$ by a function f, where $\overline{K}$ is the merge of all the DRSs to which K is subordinate. Then we may add $K'_2$ to K, where $K'_2$ results from $K_2$ by replacing all occurrences of discourse referents of $U_{K_2}$ by new ones and the discourse referents **x** declared in $U_{K_1}$ by f(**x**).

We will generalize DET to UDRSs such that the structure that results from an application of DET to a UDRS is again a UDRS, i.e. directly represents some natural language sentence. We, therefore, incorporate the task of what is usually done by a rule of thinning into the formulation of DET itself and also into the following definition of embedding. We define an *embedding* f of a UDRS into a UDRS to be a function that maps labels to labels and discourse referents to discourse referents while preserving all conditions in which they occur. We assume that f is one-to-one when f is restricted to the set of discour-se referents occurring in proper sub-universes. Only discourse referents occurring in the universe associated with $l_\top$ may be identified by f. We do not assume that the restriction of f to the set of labels is one-to-one also. But f must preserve $\neg$, $\Rightarrow$ and $\vee$, i.e. respect the following restrictions.

---

[4] We define $l < l' := l \leq l' \land l \neq l'$.

[5] For the construction of underspecified representations see [2], this volume.

[6] We will use implicative conditions of the form $\langle \Rightarrow, l_1, l_2 \rangle$, to represent universally quantified NPs (instead of their generalized quantifier representation $\langle every, l_1, l_2 \rangle$).

(i) if $l{:}{\Rightarrow}(l_1,l_2)$ occurs in $K'$, then $f(l){:}{\Rightarrow}(f(l_1),f(l_2))$,
(ii) if $l{:}\neg l_1$ occurs in $K'$, then $f(l){:}\neg f(l_1)$.

For the formulation of the deduction rules it is convenient to introduce the following abbreviation. Let $\mathcal{K}$ be a UDRS and $l$ some of its labels. Then $\mathcal{K}_l$ is the sub-UDRS of $\mathcal{K}$ dominated by $l$, i.e. $K_l$ contains all conditions $l'{:}\gamma$ such that $l'\leq l$ and its ordering relation is the restriction of $\mathcal{K}$'s ordering relation.

Suppose $\gamma = l_0{:}l_1{\Rightarrow}l_2$ is the distinguished condition of a UDRS component $l{:}K$ occurring in a UDRS clause $\mathcal{K}_l$ of a UDRS $\mathcal{K}$. And suppose there is an embedding f of $\mathcal{K}_{l_1}$ into a set of conditions $l'{:}\delta$ of $\mathcal{K}$ such that $l \leq l'$. Then the result of an application of DET to $\gamma$ is a clause $\mathcal{K}'_l$ that is obtained from $\mathcal{K}_l$ by (i) eliminating $\mathcal{K}_{l_1}$ from $\mathcal{K}_l$ (ii) replacing all occurrences of discourse referents in the remaining structure by new ones and the discourse referents $x$ declared in the universe of $l_1$, by $f(x)$; (iii) substituting $l'$ for $l$, $l_1$, and $l_2$ in $ORD_l$; and (iv) replacing all other labels of $\mathcal{K}_l$ by new ones.

But note that applications of DET are restricted to NPs that occur 'in the context of' implicative conditions, or monotone increasing quantifiers, as shown in (16). Suppose we know that John is a politician, then:

(16) Few problems preoccupy every politician.
  $\not\vdash$ Few problems preoccupy John.
  Every politician didn't sleep.
  $\not\vdash$ John didn't sleep.
  At least one problem preoccupies every pol.
  $\vdash$ At least one problem preoccupies John.

(16) shows that DET may only be applied to a condition $\gamma$ occurring in $l{:}K$, if there is no component $l'{:}K'$ such that the distinguished condition $l'{:}\gamma'$ of $K'$ is either a monotone decreasing quantifier or a negation, and such that for some disambiguation of the clause in which $\gamma$ occurs we get $l \leq scope(l')$. As the negation of a monotone decreasing quantifier is monotone increasing and two negations neutralize each other the easiest way to implement the restriction is to assign polarities to UDRS components and restrict applications of DET to components with positive polarity as follows.

Suppose $l{:}K$ occurs in a UDRS clause
$\langle l_0{:}\langle\gamma_0,...,\gamma_n\rangle, ORD_{l_0}\rangle$, where $l_0$ has positive polarity, written $l_0^+$. Then $l$ has *positive (negative) polarity* if for each disambiguation the cardinality of the set of monotone decreasing components (i.e. monotone decreasing quantifiers or negations) that takes wide scope over $l$ is even (odd). Negative polarity of $l_0$ is induces the complementary distribution of polarity marking for $l$. If $l$ is the label of a complex condition, then the polarity of $l$ determines the polarity of the arguments of this condition according to the following patterns: $l^+{:}l_1^-\Rightarrow l_2^+$, $l^-{:}l_1^+\Rightarrow l_2^-$, $l^+{:}\neg l_1^-$, and $l^-{:}\neg l_1^+$. $l_\top^i$ has positive polarity for every $i$. The polarity of the upper bound label of a UDRS-clause is inherited from the polarity of the label the UDRS-clause is attached to. Verbs, i.e. lower bounds of UDRS-clauses, always have definite polarities if the upper bound label of the same clause has.

Two remarks are in order before we come to the formulation of DET. First, the polarity distribution can be done without explicitly calculating all disambiguations. The label $l$ of a component $l{:}K$ is positive (negative) in the clause in which occurs, if the set of components on the path to the upper bound label $l_0^+$ of this clause contains an even (odd) number of polarity changing elements, and all other components of the clause (i.e. those occurring on other paths) do not change polarity. Second, the fragment of UDRSs we are considering in this paper does not contain a treatment of n-ary quantifiers. Especially we do not deal with resumptive quantifiers, like $<no\ boy, no\ girl>$ in **No boy likes no girl**. If we do not consider the fact that this sentence may be read as **No boy likes any girl** the polarity marking defined above will mark the label of the verb as positive. But if we take this reading into account, i.e. allow to construe the two quantified NPs as constituents of the resumptive quantifier, then one negation is cancelled and the label of the verb cannot get a definite value.[7]

To represent DET schematically we write $\langle l_\top{:}\sigma(l^\pi{:}\gamma), \mathrm{ORD}\rangle$ to indicate that $l^\pi{:}K$ is a component of the UDRS $\mathcal{K}_{l_\top}$ with polarity $\pi$ and distinguished condition $\gamma$.

$$\frac{\Delta \quad \langle l_\top{:}\sigma(l_i^+{:}l_{i1}^- \Rightarrow l_{i2}^+), \mathrm{ORD}\rangle \quad f: \mathcal{K}_{l_{i1}} \mapsto \Delta\ \text{exists}}{f(\mathcal{K}_{l_{i2}})}$$

The scheme for DET allows the arguments of the implicative condition to which it is applied still to be ambiguous. The discussion of example (6) in Section 2 focussed on the ambiguity of its antecedent only. (We ignored the ambiguity of the consequent there.) To discuss the case of ambiguous consequents we consider the the following argument.

(17) If the chairman talks, everybody doesn't sleep.
  The chairman talks. $\vdash$ Everybody doesn't sleep.

There is a crucial difference between (17) and (6): The truth of the conclusion in (17) depends on the fact that it is derived from the conditional. It, therefore, must be treated as correlated with the consequent of the conditional under any disambiguation. No non-correlated disambiguations are allowed. To ensure this we must have some means to represent the 'history' of the clauses that are added to a set of data. As (8) suggests this could be done by coindexing $\mathcal{K}_{l_{i1}}$ and $\mathcal{K}_{f(l_{i1})}$ in the representation of (17).

---

[7] A general treatment of n-ary quantification within the theory of UDRSs has still to be worked out. In [6] it is shown how cumulative quantification may be treated using identification of labels.

In contrast to the obligatory coindexing in the case of (17) the consequence relation in (8) does allow for non-correlated interpretations in the case of (2). Such interpretations naturally occur if, e.g., the conditional and the minor premiss were introduced by very distinct parts of a text from which the database had been constructed. In such cases the interpreter may assume that the contexts in which the two sentences occurred are independent of each other. He, therefore, leaves leeway for the possibility that (later on) each context could be provided with more information in such a way that those interpretations trigger different disambiguations of the two occurrences. In such cases "crossed interpretations" must be allowed, and any application of DET must be refused by contraindexing – except the crossed interpretations can be shown to be equivalent. For the sake of readability we present the rule only for the propositional case.

$$\frac{\Delta\ \alpha_i \Rightarrow \beta_j\ \alpha_k \quad i = k \vee (i \neq k \wedge \Delta \vdash \alpha_i \Leftrightarrow \alpha_k)}{\beta_j}$$

But the interpreter could also adopt the strategy to accept the argument also in case of non-correlated interpretations *without* checking the validity of $\alpha_i \Leftrightarrow \alpha_k$. In this case he will conclude that $\beta_j$ holds under the proviso that he might revise this inference if there will be additional information that forces him to disambiguate in a non-correlated way. If then $\alpha_i \Leftrightarrow \alpha_k$ does not hold he must be able to give up the conclusion $\beta_j$ and every other argument that was based on it. To accomodate this strategy we need more than just coindexing. We need means to represent the structure of whole proofs. As we have labels available in our language we may do this by adopting the techniques of labelled deductive systems ([3]). For reasons of space we will not go into this in further detail.

The next inference rule, AI, allows one to introduce ambiguities. It contrasts with the standard rule of disjunction introduction in that it allows for the introduction of a UDRS $\sigma$ that is underspecified with respect to the two readings $\sigma_1$ and $\sigma_2$ only if both, $\sigma_1$ and $\sigma_2$, are contained in the data. This shows once more that ambiguities are not treated as disjunctions.

**Ambiguitiy Introduction** Let $\sigma_1$ and $\sigma_2$ be two UDRSs of $\Delta$ that differ only w.r.t. their ORDs. Then we may add a UDRS $\sigma_3$ to $\Delta$ that is like $\sigma_1$ but has the intersection of ORD and ORD' as ordering of its labels. The index of $\sigma_3$ is new to $\Delta$.

We give an example to show how AI and DET interact in the case of non-correlated readings: Suppose the data $\Delta$ consists of $\sigma_1$, $\sigma_2$ and $\sigma_3 \Rightarrow \gamma$. We want to derive $\gamma$. We apply AI to $\sigma_1$ and $\sigma_2$ and add $\sigma_3$ to $\Delta$. As the index of $\sigma_3$ is new we must check whether $\sigma_1 \Leftrightarrow \sigma_2$ can be derived from $\Delta$. Because $\Delta$ contains both of them the proof succeeds.

The last rule of inference, DIFF, eliminates ambiguities on the basis of structural differences in the ordering relations. Suppose $\alpha_1$ and $\alpha_2$ are a underspecified representations with three scope bearing components $l_1$, $l_2$, and $l_3$. Assume further that $\alpha_1$ has readings that correspond to the following orders of these components: $\langle l_3, l_2, l_1 \rangle$, $\langle l_2, l_3, l_1 \rangle$, and $\langle l_2, l_1, l_3 \rangle$, whereas $\alpha_2$ is ambiguous between $\langle l_2, l_3, l_1 \rangle$ and $\langle l_2, l_1, l_3 \rangle$. Suppose now that the data contains $\alpha_1$ and the negation of $\alpha_2$. Then this set of data is equivalent to the reading given by $\langle l_3, l_2, l_1 \rangle$. To see that this holds the *structural difference* between the structures $\text{ORD}_{\alpha_1}$ and $\text{ORD}_{\alpha_2}$ has to be calculated. The *structural difference* between two structures $\text{ORD}_{\alpha_1}$ and $\text{ORD}_{\alpha_2}$ is the partial order that satisfies $\text{ORD}_{\alpha_1}$ but not $\text{ORD}_{\alpha_2}$, if there is any; and it is falsity if there is no such order. Thus the notion of structural difference generalizes the traditional notion of inconsistency. Again a precise formulation of DIFF is given in [7].

## 5 Rules of Proof

Rules of proof are deduction rules that allow us to reduce the complexity of the goal by accomplishing a subproof. We will consider COND(itionalization) and R(eductio)A(d)A(bsurdum) and show that they may not be applied in the case of ambiguous goals (i.e. goals in which no operator has widest scope).

Suppose we want to derive **everybody didn't snore** from **everybody didn't sleep** and the fact that snoring implies sleeping. I.e. we want to carry out the proof in (18), where ORD = $\{l_3 \leq scope(l_1), l_3 \leq scope(l_2), l_5 \leq scope(l_4)\}$ and ORD' = $\{l_8 \leq scope(l_7), l_8 \leq scope(l_6)\}$.

$$\langle l_\top : \langle l_1 : \boxed{\begin{array}{c} x \\ \phantom{x} \end{array}} \Rightarrow \boxed{\phantom{x}}, l_2 : \neg \boxed{\phantom{x}}, l_3 : \boxed{\text{x sleep}} \rangle, ORD \rangle$$

$$\langle l_\top : \langle l_4 : \boxed{\begin{array}{c} x \\ \text{x snore} \end{array}} \Rightarrow \boxed{\phantom{x}}, l_5 : \boxed{\text{x sleep}} \rangle, ORD \rangle$$

$$\overline{\langle l'_\top : \langle l_6 : \boxed{\begin{array}{c} x \\ \phantom{x} \end{array}} \Rightarrow \boxed{\phantom{x}}, l_7 : \neg \boxed{\phantom{x}}, l_8 : \boxed{\text{x snore}} \rangle, ORD' \rangle}$$

(18)

Let us try to apply rules of proof to reduce the complexity of the goal. We use the extensions of COND and RAA given in [7]. There use is quite simple. An application of COND to the goal in (18) results in adding $\langle l_\top : \boxed{a}, \{\ \} \rangle$ to the data and leaves $\langle l'_\top : \langle l_7 : \neg \boxed{a}, l_8 : \boxed{\text{a snore}} \rangle, \text{ORD"} \rangle$ to be shown, where $ORD''$ results from $ORD'$ by replacing $l_6$ and $scope(l_6)$ with $l'_\top$. RAA is now applicable to the new goal in a standard way. It should be clear, however, that the order of application we have cho-sen, i.e. COND before RAA, results in having given the universal quantifier wide scope over the negation. This means that after having applied COND we are not in the process of proving the original ambiguous goal any more. What we are going to prove instead is that reading of the goal with universal quantifier having wide scope over the negation. Beginning with RAA instead of COND assigns the negation wide scope over the quantifier, as we would add $\langle l'_\top : \langle l_6 : \boxed{x} \Rightarrow \boxed{}, l_8 : \boxed{x \text{ snore}} \rangle, ORD'' \rangle$ to the data in order to derive a contradiction.[8] Here $ORD''$ results from $ORD'$ by replacing $l_7$ and $scope(l_7)$ with $l'_\top$.

If we tried to keep the reduction-of-the-goal strategy we would have to perform the disambiguation steps to formulas in the data that the order of application on COND and RAA triggers. And in addition we would have to check all possible orders, not only one. Hence we would perform exactly the same set of proofs that would be needed if we represented ambiguous sentences by sets of formulas. Nothing would have been gained with respect to any traditional approach.

We thus conclude that applications of COND and RAA are *only* possible if either $\Rightarrow$ or $\neg$ has wide scope in the goal. In this case standard formulations of COND and RAA may be applied even if the goal is ambiguous at some lower level of structure. In case the underspecification occurs with respect to the relative scope of immediate daughters of $l_\top$, however, we must find some other means to relate non-identical UDRSs in goal and data. What we need are rules for UDRSs that generalize the success case for atoms within ordinary deduction systems.

## 6 Deduction rules for top-level ambiguities

The inference in (18) can be realised very easily if we allow components of UDRSs that are marked negative to be replaced by components with a smaller denotation. Likewise components of UDRSs that are marked positive may be replaced by components with a larger denotation. If the component to be replaced is the restrictor of a generalized quantifier, then in addition to the polarity marking the soundness of such substitutions depends on the persistence property of the quantifier. In the framework of UDRSs persistence of quantifiers has to be defined relative to the context in which they occur. Let $NP_i$ be a persistent (anti-persistent) NP. Then $NP_i$ is called *persistent (anti-persistent) in clause* S, if this property is preserved under each disambiguation of S. So **everybody** is anti-persistent in (19e), but not in (19a), because the wide scope reading for the negation blocks the inference in (19b). It is not persistent in (19c) nor in (19d).

(19) a. Everybody didn't come.
 b. Everybody didn't come.
   $\not\models$ Every woman didn't come.
 c. More than half the problems were solved by everybody.
 d. It is not true that everybody didn't come.
 e. Some problem was solved by everybody.

The main rule of inference for UDRSs is the following R(eplacement)R(ule).

**RR** Whenever some UDRS $\mathcal{K}^i_\top$ occurs in a UDRS-database $\Delta$ and $\Delta \vdash \mathcal{K}^i_\top \gg \mathcal{K}'^i_\top$ holds, then $\mathcal{K}'^i_\top$ may be added to $\Delta$.

RR is based on the following substitution rule. The $\gg$-rules are given below.

**SUBST** Let $l:K$ be a DRS component occurring in some UDRS $\mathcal{K}$, $\Delta$ a UDRS-database. Let $\mathcal{K}'$ be the UDRS that results from $\mathcal{K}$ by substituting $K'$ for $K$.
Then $\Delta \vdash \mathcal{K} \gg \mathcal{K}'$, if (i) or (ii) holds.
 (i) $l$ has positive polarity and $\Delta \vdash K \gg K'$.
 (ii) $l$ has negative polarity and $\Delta \vdash K' \gg K$.

Schematically we represent the rule (for the case of positive polarity) as follows.

$$\frac{\Delta, \mathcal{K}^i_\top \hookleftarrow l^+:K}{\Delta, \mathcal{K}^i_\top \hookleftarrow l^+:K'} \quad \text{if} \quad \Delta \vdash l^+:K \gg l^+:K'$$

For UDRS-components we have the following rule.

$\gg$ **DRS:** $\Delta \vdash K \gg K'$ if there is a function $f: U_K \to U_{K'}$ such that for all $\gamma' \in C_{K'}$ there is a $\gamma \in C_K$ with $\Delta \vdash f(\gamma) \gg \gamma'$.[9]

Complex conditions are dealt with by the following set of rules. Except for persistence properties they are still independent of the meaning of any particular generalized quantifier. The success of the rules can be achieved in two ways. Either by recursively applying the $\gg$-rules. Or, by proving the implicative condition which will guarantee soundness of SUBST.

$\gg \Rightarrow$:
 $\Delta \vdash \langle \Rightarrow, l_1, l_2 \rangle \gg \langle \Rightarrow, l'_1, l'_2 \rangle$ if
 1. $\Delta \vdash \mathcal{K}_{l_1} \gg \mathcal{K}_{l'_1}$, or
 2. $\Delta \vdash \langle \to, \mathcal{K}_{l_1}, \mathcal{K}_{l'_1} \rangle$

$\gg Q$:
 (i) $\Delta \vdash \langle Q, l_1, l_2 \rangle \gg \langle Q, l'_1, l'_2 \rangle$ if $Q$ is persistent and
  1. $\Delta \vdash \mathcal{K}_{l_1} \gg \mathcal{K}_{l'_1}$, or
  2. $\Delta \vdash \langle \to, \mathcal{K}_{l_1}, \mathcal{K}_{l'_1} \rangle$

 (ii) $\Delta \vdash \langle Q, l_1, l_2 \rangle \gg \langle Q, l'_1, l'_2 \rangle$ if $Q$ is anti-pers. and
  1. $\Delta \vdash \mathcal{K}_{l'_1} \gg \mathcal{K}_{l_1}$, or

---

[8] If we would treat ambiguous clauses as the disjunctions of their meanings, i.e. take the consequence relation in (1), then this disambiguation could be compensated for by applying RESTART (see [7] for details). But relative to the consequence relation under (8) RESTART is not sound!

[9] $f(\gamma)$ is $\gamma$ with discourse referents $x$ occurring in $\gamma$ replaced by $f(x)$.

2. $\Delta \vdash \langle \rightarrow, \mathcal{K}_{l'_1}, \mathcal{K}_{l_1} \rangle$

$\gg \neg$:
$\Delta \vdash \langle \neg, l_1 \rangle \gg \langle \neg, l'_1 \rangle$ if
1. $\Delta \vdash \mathcal{K}_{l'_1} \gg \mathcal{K}_{l_1}$, or
2. $\Delta \vdash \langle \rightarrow, \mathcal{K}_{l'_1}, \mathcal{K}_{l_1} \rangle$

The following rules involve lexical meaning of words. We give some examples of determiner rules to indicate how we may deal with the logic of quantifiers in this rule set. Rules for nouns and verbs refer to a further inference relation, $\vdash^{\mathcal{L}}$. This relation takes the meaning postulates into account that a particular lexical theory associates with particular word meanings.

$\gg$ *Lex*:
(i) $\langle every, l_1, l_2 \rangle \gg \langle more\ than\ half, l_1, l_2 \rangle$
(ii) $\langle every, l_1, l_2 \rangle \gg \langle \{\}, \{Mary\}, l_2 \rangle$
(iii) $\langle no, l_1, l_2 \rangle \gg \langle every, l_1, l'_2 : \neg l_2 \rangle$
(iv) $\langle some, l_1, l'_2 : \neg l_2 \rangle \gg \langle not\ every, l_1, l_2 \rangle$
(v) $snore \gg sleep$ if $\vdash^{\mathcal{L}} snore \gg sleep$

The last rule allows relative scopes of quantifiers to be inverted.

$\gg \pi$:
(i) Let $l_1^+ : \gamma_1$ and $l_2 : \gamma_2$ be two quantifiers of a UDRS $\mathcal{K}$ such that $l_1$ immediately dominates $l_2$ ($l_2 \leq^i scope(l_1)$). Let $\pi$ be the relation between quantifiers that allows neigbourhood exchanges, i.e. $\gamma_1 \pi \gamma_2$ iff $\mathcal{K}_{l_1} \vdash \mathcal{K}'_{l_1}$, where $\mathcal{K}'_{l_1}$ results from $\mathcal{K}_{l_1}$ by exchanging $\gamma_1$ and $\gamma_2$, i.e. by replacing $l_2 \leq^i scope(l_1)$ in $\mathcal{K}_{l_1}$'s $ORD$ by $l_1 \leq^i scope(l_2)$. Then
$\Delta \vdash \mathcal{K}_{l_1} \gg \mathcal{K}'_{l_1}$ if $l_1 : \gamma_1 \pi l_2 : \gamma_2$ and $l_1 : \gamma_1 \pi l' : \gamma'$ for all $l' : \gamma'$ that may be immediately dominated by $l_1 : \gamma_1$ (in any disambiguation).
(ii) Analoguously for the case of $l_1^- : \gamma_1$ having negative polarity.

The formulation of this rule is very general. In the simplest case it allows one to derive a sentence where an indefinite quantifier is interpreted non-specifically from an interpretation where it is assigned a specific meaning. If the specific/non-specific distinction is due to a universally quantified NP then the rule uses the fact that $\langle a, l, s \rangle \pi \langle every, l, s \rangle$ holds. As other scope bearing elements may end up between the indefinite and the universal in some disambiguation the rule may only be applied, if these elements behave exactly the same way as the universal does, i.e. allow the indefinite to be read non-specifically. In case such an element is another universally quantified NP we thus may apply the rule, but we cannot apply it is a negation.

## 7 Conclusion and Further Perspectives

The paper has shown that it is possible to reason with ambiguities in a natural, direct and intuitively correct way.

The fact that humans are able to reason with ambiguities led to a natural distinction between deduction systems that apply rules of proof to reduce the complexity of a goal and systems of logic that are tailored directly for natural language interpretation and reasoning. Human interpreters seem to use both systems when they perform reasoning tasks. We know that we cannot surmount undecidability (in a non-adhoc way) if we take quantifiers and/or connectives as logical devices in the traditional sense. But as the deduction rules for top-level ambiguities given here present an extension of Aristotelian syllogism metamathematical results about their complexity will be of great interest as well as the proof of a completeness theorem. Apart from this research the use of the rule system within the task of natural language understanding is under investigation. It seems that the Replacement Rules are particularly suited to do special reasoning tasks necessary to disambiguate lexical ambiguities, because most of the deductive processes needed there are independent of any quantificational structure of the sentences containing the ambiguous item.

## Acknowledgements

The ideas of this paper where presented, first at an international workshop of the SFB 340 "Sprachtheoretische Grundlagen der Computerlinguistik" in October 1993, and second, at a workshop on 'Deduction and Language' that took place at SOAS, London, in spring 1994. I am particularly grateful for comments made by participants of these workshops.

## Literatur

[1] Hiyan Alshawi and Richard Crouch. Monotonic semantic interpretation. In *Proceedings of ACL*, pages 32–39, Newark, Delaware, 1992.

[2] Anette Frank and Uwe Reyle. Principle based semantics for hpsg. In *Proceedings of EACL 95, Dublin*, 1995.

[3] Dov Gabbay. Labelled deductive systems. Technical report, Max Planck Institut für Informatik, 1994.

[4] Hans Kamp and Uwe Reyle. Technical report.

[5] Massimo Poesio. Scope ambiguity and inference. Technical report, University of Rochester, N.Y., 1991.

[6] Uwe Reyle. Monotonic disambiguation and plural pronoun resolution. In Kees van Deemter and Stanley Peters, editors, *CSLI Lecture Notes: Semantic Ambiguity and Underspecification*.

[7] Uwe Reyle. Dealing with ambiguities by underspecification: Construction, representation, and deduction. *Journal of Semantics*, 10(2), 1993.

[8] Kees van Deemter. *On the Composisiton of Meaning*. PhD thesis, University of Amsterdam, 1991.